\documentclass[10pt,twocolumn,letterpaper]{article}

\usepackage{iccv}
\usepackage{times}
\usepackage{epsfig}
\usepackage{graphicx}
\usepackage{amsmath}
\usepackage{amssymb}
\usepackage{url}
\usepackage{multirow}
\usepackage{adjustbox}
\usepackage{threeparttable}
\usepackage{algorithm}
\usepackage{setspace}
\usepackage[table,xcdraw]{xcolor}
\usepackage{amsmath}
\usepackage{amssymb}
\usepackage{multirow}
\usepackage{adjustbox}
\usepackage{threeparttable}
\usepackage{algorithmic}

\usepackage[breaklinks=true,bookmarks=false]{hyperref}
\hypersetup{
	linkcolor=red,
	filecolor=red,
	citecolor=green,      
	urlcolor=magenta,
}

\iccvfinalcopy 

\def\iccvPaperID{3832} 

\ificcvfinal\fi

\begin{document}

\title{\noindent\rule[0.25\baselineskip]{\textwidth}{2.5pt} \\ \LARGE ActionCLIP: A New Paradigm for Video Action Recognition \\
	\noindent\rule[0.25\baselineskip]{\textwidth}{1.5pt}
}
\author{
	Mengmeng Wang\\ 
	Zhejiang University\\
	{\tt\small mengmengwang@zju.edu.cn}
	\and
	Jiazheng Xing\\
	Zhejiang University\\
	{\tt\small jiazhengxing@zju.edu.cn}
	\and
	Yong Liu\thanks{Corresponding author.}\\
	Zhejiang University\\
	{\tt\small yongliu@iipc.zju.edu.cn}
}

\maketitle
\ificcvfinal\thispagestyle{empty}\fi

\begin{abstract}
	\begin{spacing}{0.95}
	\small The canonical approach to video action recognition dictates a neural model to do a classic and standard 1-of-N majority vote task. They are trained to predict a fixed set of predefined categories, limiting their transferable ability on new datasets with unseen concepts. In this paper, we provide a new perspective on action recognition by attaching importance to the semantic information of label texts rather than simply mapping them into numbers. Specifically, we model this task as a video-text matching problem within a multimodal learning framework, which strengthens the video representation with more semantic language supervision and enables our model to do zero-shot action recognition without any further labeled data or parameters requirements. Moreover, to handle the deficiency of label texts and make use of tremendous web data, we propose a new paradigm based on this multimodal learning framework for action recognition, which we dub \textit{``pre-train, prompt and fine-tune"}. This paradigm first learns powerful representations from pre-training on a large amount of web image-text or video-text data. Then it makes the action recognition task to act more like pre-training problems via prompt engineering. Finally, it end-to-end fine-tunes on target datasets to obtain strong performance. We give an instantiation of the new paradigm, \textbf{ActionCLIP}, which not only has superior and flexible zero-shot/few-shot transfer ability but also reaches a top performance on general action recognition task, achieving 83.8\% top-1 accuracy on Kinetics-400 with a ViT-B/16 as the backbone. 
	Code is available at \url{https://github.com/sallymmx/ActionCLIP.git}.
\end{spacing}
\end{abstract}

\section{Introduction}
\begin{figure} [ht]
	\centering
	\includegraphics[width=\linewidth,height=0.75\linewidth]{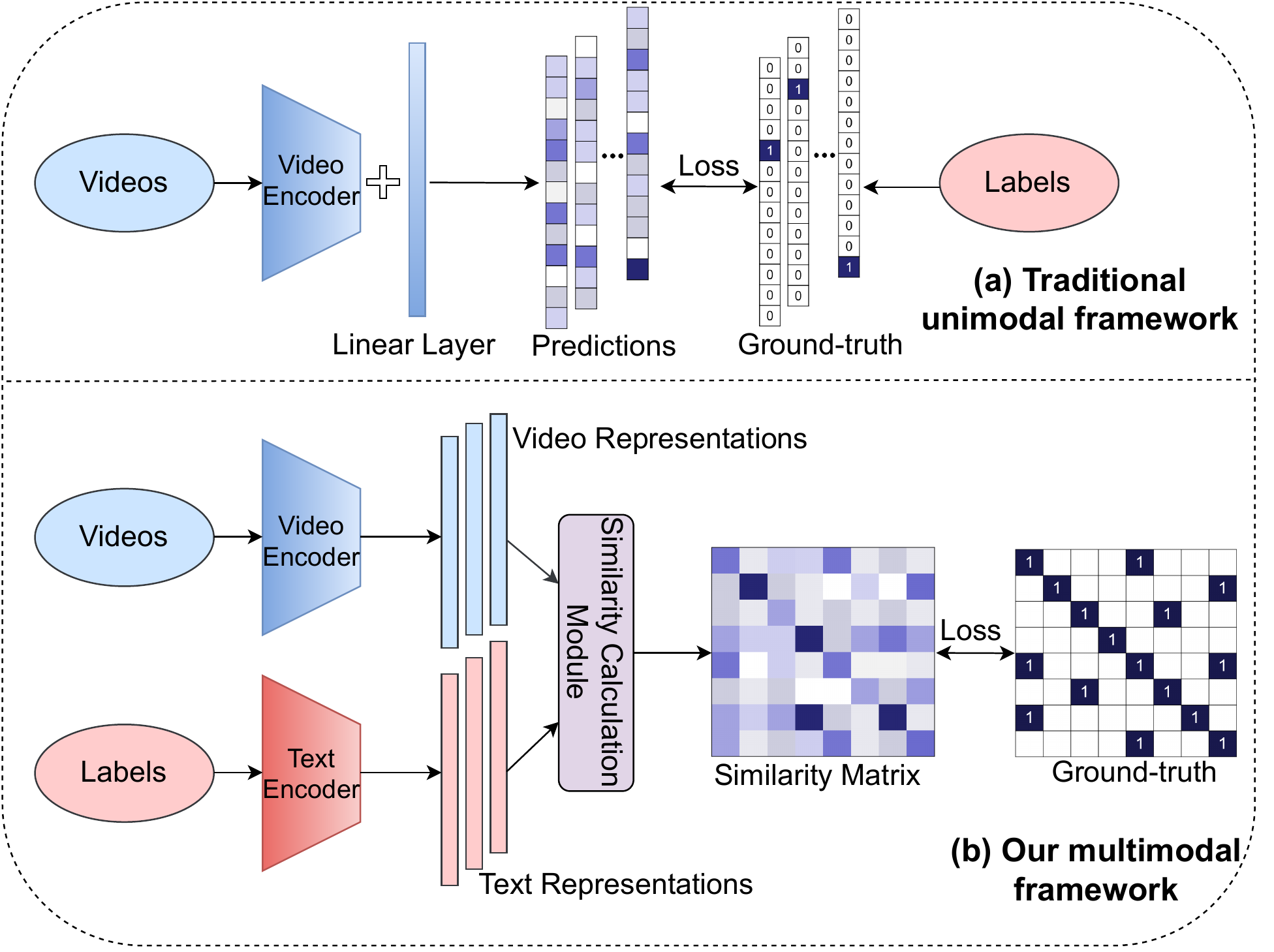}
	\caption{Existing unimodality pipeline (a) and our multimodal framework (b). They are different in the usage of labels. (a) maps labels into numbers or one-hot vectors while (b) exploits the semantic information of label text itself and tries to pull the corresponding video representation close to each other.}
	\label{fig:fig1}
\end{figure}
Video action recognition is the first step of video understanding, and it is an active research area in recent years. We have observed that it mainly went through two stages, \textit{feature engineering} and \textit{architecture engineering}. Since there were no sufficient data for learning high-quality models before the birth of large datasets like Kinetics~\cite{carreira2017quo}, early methods focused on the \textit{feature engineering}, where researchers considered the temporal information inside the videos and used their knowledge to design specific hand-crafted representations~\cite{dollar2005behavior,wang2013dense}. Then, with the advent of deep neural networks and large benchmarks, we are now in the second stage, \textit{architecture engineering}. Lots of well-designed networks sprang up by reasonably absorbing the temporal dimension like two-stream networks~\cite{wang2016temporal}, 3D convolutional neural networks (CNN)~\cite{feichtenhofer2019slowfast}, compute-efficient networks~\cite{jiang2019stm} and transformer-based networks~\cite{arnab2021vivit}. 

Though the features and network architectures have been well-studied in the last few years, they are trained to predict a fixed set of predefined categories within a unimodal framework as shown in Figure \ref{fig:fig1}(a). This predetermined manner limits their generality and employment since additional labeled training data is required to transfer to any other new and unseen concepts. Instead of directly mapping labels to numbers like traditional works, learning from the raw text will be a promising solution which could be a much broader source of supervision and provide a more comprehensive representation. Reminiscent of how our humans do this job, we can recognize both known and unknown videos by associating the semantic information from the visual appearance to natural language sources rather than numbers. In this paper, we explore the natural language supervision in a multimodal framework as shown Figure \ref{fig:fig1}(b) with two objectives, i) strengthening the representation of the traditional action recognition with more semantic language supervision, and ii) enabling our model to realize zero-shot transfer without any further labeled data or parameters requirements. Our multimodal framework includes two separate unimodal encoders for videos and labels and a similarity calculation module. The training objective is to pull the pairwise video and label representations close to each other, thus the learned representations will be more semantic than unimodal methods. In the inference phase, it becomes a video-text matching problem rather than a classical 1-of-N majority vote task and is capable of zero-shot prediction.

However, labels of existing fully-supervised action recognition datasets are always too succinct to construct rich sentences for language learning. Collecting and annotating new video datasets require huge storage resources and enormous human effort and time. On the other hand, a sea of videos with noisy but rich text labels are stored and generated on the web every day. Is there a way to energize the abundant web data for action recognition?  Pre-training may be a solution that is demonstrated in ViViT~\cite{arnab2021vivit}. But it is not easy to pre-train with a large magnitude of web data. It is expensive on storage hardware, computational resources and experiment cycles$\footnote{\cite{dosovitskiy2020image} reports that pre-training a ViT-H/14 model on JFT takes 2.5k TPUv3-core-days}$. This triggers another motivation of this paper, could we directly adapt a pre-trained multimodal model into this task, avoiding the above dilemma? We find this is possible. Formally, we define a new paradigm ``\textit{pre-train, prompt, and fine-tune}" for video action recognition. Although it is appealing to \textit{pre-train} the whole
model end-to-end with large-scale video-text datasets such
as HowTo100M~\cite{miech2019howto100m}, we are restricted by the enormous computation cost. Luckily, we find it is also worked to use a pre-trained model. Here we use the word ``\textit{pre-train}" rather than ``\textit{pre-trained}" in the new paradigm to keep the pre-training function. Then, instead of adapting the pre-trained model in specific benchmarks by substituting the final classification layers and objective functions, we reformulate our task to look more like those solved during the original pre-training procedure via \textit{prompt}. Prompt-based learning~\cite{liu2021pre} is regarded as a sea change to natural language processing (NLP), but it is not active in vision tasks, especially has not been exploited in action recognition. We believe it will have attractive prospects in many vision-text-related tasks and explore it in action recognition here. Finally, we \textit{fine-tune} the whole model on target datasets. We implement an instantiation of this paradigm, \textbf{ActionCLIP}, which employs CLIP~\cite{radford2021learning} as the pre-trained model. It obtains a top performance of 83.8\% top-1 accuracy on Kinetics-400. Our contributions can be summarized as follows:
\begin{itemize}
	\item We formulate the action recognition task as a multimodal learning problem rather than a traditional unimodal classification task. It strengthens the representations with more semantic language supervision and enlarges the generality and employment of the model in zero-shot/few-shot situations.  
	
	\item We propose a new paradigm for action recognition, which we dub ``\textit{pre-train, prompt, and fine-tune}". In this paradigm, we could directly reuse powerful large-scale web data pre-trained models by designing appropriate prompts, significantly reducing the pre-training cost. 
	\item Comprehensive experiments demonstrate the potential and effectiveness of our method, which consistently outperforms the state-of-the-art methods on several public benchmark datasets. 
	
\end{itemize}

\section{Related Works}
\subsection{Video Action Recognition}
We have observed that video action recognition mainly went through two stages, \textit{feature engineering} and \textit{architecture engineering}. In the first stage, lots of hand-craft descriptors are designed for spatio-temporal representations, like Cuboids~\cite{dollar2005behavior}, 3D 3DHOG~\cite{klaser2008spatio} and Dense Trajectories~\cite{wang2013dense}. However, these features lack generalization since they are not end-to-end learned in large-scale datasets. Now we are in the second stage, \textit{architecture engineering}. We coarsely classify these architectures into four categories, two-stream networks, 3D CNNs, compute-efficient networks and transformer-based networks. Two-stream-based methods \cite{feichtenhofer2016convolutional, wang2016temporal, wang2017spatiotemporal} are introduced to model appearance and dynamics separately with two networks and fuse two streams through the middle or at last. 3D CNNs \cite{carreira2017quo,diba2018spatio,feichtenhofer2019slowfast, stroud2020d3d} intuitively learn spatiotemporal features from RGB frames directly which extend the common 2D CNNs with an extra temporal dimension. Due to the heavy computational burden of 3D CNNs, many compute-efficient networks are designed to find the trade-off between precision and speed \cite{tran2018closer,xie2018rethinking,zhou2018mict,jiang2019stm,kumawat2021depthwise,li2020tea}. Transformer-based networks~\cite{arnab2021vivit,bertasius2021space,neimark2021video,sharir2021image,fan2021multiscale} employ and modify recent strong vision transformers to jointly encode the spatial and temporal features. Yet, most works of both stages are unimodal, without considering the semantic information contained in the labels. We propose a new paradigm ``\textit{pre-trained, prompt, and fine-tune}" based on a video-text multimodal learning framework for action recognition, which sheds light on the language modeling of label words. 

\subsection{Vision-text Multi-modality in Action Recognition}
Vision-text multi-modality is a hot topic in several vision-text related fields recently, like
pre-training~\cite{lei2021less,li2021align}, vision-text retrieval~\cite{fang2021clip2video,miech2018learning} and so on. Video action recognition could be interpreted as a text-insufficient video-to-text retrieval problem. Therefore, it may also be feasible to apply vision-text multi-modal learning in this task. However, to the best of our knowledge, we have not found mature and effective methods from this perspective in general video action recognition. We do find several vision-text multi-modality works in self-supervised video representation learning~\cite{miech2020end,alayrac2020self} and zero-shot action recognition~\cite{piergiovanni2020learning,brattoli2020rethinking,zhang2018cross}. Yet, the former is prone to just learn a strong pre-trained video representation with a large web dataset and still neglects the label texts features when doing specific classification, just attaching and learning a linear classifier on top of the learned vision representation. The latter mainly concentrates on the embedding space designation with a pre-trained vision model and a simple text embedding like Word2Vec~\cite{mikolov2013efficient}, paying less attention to the upstream general action recognition task. Different from them, in this paper, we focus on the vision-text multi-modality learning in general action recognition and build a bridge for it and zero-shot/few-shot action recognition.
%

\section{Method}

\subsection{Multimodal Learning Framework}
Previous comprehensive video action recognition methods treat this task as a classic and standard 1-of-N majority vote problem, mapping labels into numbers. This pipeline completely ignores the semantic information contained in the label texts. We instead model this task as a video-text multimodal learning problem, in contrast to pure video modeling. We believe learning from the supervision of natural language could not only enhance the representation power but also enable flexible zero-shot transfer.

Formally, given an input video $\textbf{\emph{x}}$ and a label $\textbf{\emph{y}}$ from a predefined label set $\mathcal{Y}$, the prior works usually train a model to predict the conditional probability $P(\textbf{\emph{y}}|\textbf{\emph{x}},\theta)$ and turn $\textbf{\emph{y}}$ into a number or a one-hot vector to indicate its index of the whole label set length $|\mathcal{Y}|$. In the inference phase, the highest-scoring index of the prediction is regarded as the corresponding category. We try to break this routine and model the problem as $P(f(\textbf{\emph{x}},\textbf{\emph{y}})|\theta)$, where $\textbf{\emph{y}}$ is the original words of the label and $f$ is a similarity function. Then, the testing is more likely a matching process, the label words of the highest similarity score is the classification result:
\begin{equation}\label{eq1}
\hat{\textbf{\emph{y}}}=\underset{\textbf{\emph{y}}\in\mathcal{Y}}{argmax}P(f(\textbf{\emph{x}},\textbf{\emph{y}})|\theta)
\end{equation} 

As shown Figure \ref{fig:fig1}(b), we learn separate unimodal encoders $g_{V}, g_{W}$ for video and label words inside a dual-stream framework. The video encoder $g_{V}$ extracts spatio-temporal features for the visual modality and could be any well-designed architectures. The language encoder $g_{W}$ is used to extract features of input label texts and could be a wide variety of language models. Then, to pull the pairwise video and label representations close to each other, we define symmetric similarities between the two modalities with cosine distances in the similarity calculation module:
\begin{equation} \label{simi}
s(\textbf{\emph{x}},\textbf{\emph{y}})=\frac{\mathbf{v}\cdot \mathbf{w}^\top}{\left \| \mathbf{v}\right \|\left \| \mathbf{w}\right \|},
s(\textbf{\emph{y}},\textbf{\emph{x}})=\frac{\mathbf{w}\cdot \mathbf{v}^\top}{\left \| \mathbf{w}\right \|\left \| \mathbf{v}\right \|}
\end{equation}
where $\mathbf{v}=g_{V}(\textbf{\emph{x}})$ and $\mathbf{w}=g_{W}(\textbf{\emph{y}})$ are encoded features of $\textbf{\emph{x}}$ and $\textbf{\emph{y}}$, respectively. Then the softmax-normalized video-to-text and text-to-video similarity scores can be calculated as:
\begin{equation}
\begin{aligned}
p_{i}^{\textbf{\emph{x}}2\textbf{\emph{y}}}(\textbf{\emph{x}})&=\frac{\mathrm{exp}(s(\textbf{\emph{x}},\emph{y}_{i})/\tau)}{\sum_{j=1}^{N}\mathrm{exp}(s(\textbf{\emph{x}},\emph{y}_{j})/\tau)},\\
p_{i}^{\textbf{\emph{y}}2\textbf{\emph{x}}}(\textbf{\emph{y}})&=\frac{\mathrm{exp}(s(\textbf{\emph{y}},\emph{x}_{i})/\tau)}{\sum_{j=1}^{N}\mathrm{exp}(s(\textbf{\emph{y}},\emph{x}_{j})/\tau)}
\end{aligned}
\end{equation}
where $\tau$ is a learnable temperature parameter and $N$ is the number of training pairs. Let $\textbf{\emph{q}}^{\textbf{\emph{x}}2\textbf{\emph{y}}}(\textbf{\emph{x}}), \textbf{\emph{q}}^{\textbf{\emph{y}}2\textbf{\emph{x}}}(\textbf{\emph{y}})$ indicate the ground-truth similarity scores, where the negative pair has a probability of 0 and the positive pair has a probability of 1. Since the amount of videos are much larger than the fixed labels, it will inevitably appear multiple videos belonging to one label in a batch. Therefore, it may exist more than one positive pair in both $q_{i}^{\textbf{\emph{x}}2\textbf{\emph{y}}}(\textbf{\emph{x}})$ and $q_{i}^{\textbf{\emph{y}}2\textbf{\emph{x}}}(\textbf{\emph{y}})$. It is not proper to regard the similarity score learning as a 1-in-N classification problem with cross-entropy loss. Instead, we define the Kullback--Leibler (KL) divergence as the video-text contrastive loss to optimize our framework as:
\begin{equation}\label{kl}
\mathcal{L}=\frac{1}{2}\mathbb{E}_{(\textbf{\emph{x}},\textbf{\emph{y}})\sim \mathcal{D}}[\mathrm{KL}(\textbf{\emph{p}}^{\textbf{\emph{x}}2\textbf{\emph{y}}}(\textbf{\emph{x}}),\textbf{\emph{q}}^{\textbf{\emph{x}}2\textbf{\emph{y}}}(\textbf{\emph{x}}))+\mathrm{KL}(\textbf{\emph{p}}^{\textbf{\emph{y}}2\textbf{\emph{x}}}(\textbf{\emph{y}}),\textbf{\emph{q}}^{\textbf{\emph{y}}2\textbf{\emph{x}}}(\textbf{\emph{y}}))]
\end{equation}
where $\mathcal{D}$ is the whole training set. Based on the multimodal framework, we can simply carry out zero-shot prediction as the normal testing process in Equation~\ref{eq1}.
\begin{figure*} [ht]
	\centering
	\includegraphics[width=\linewidth]{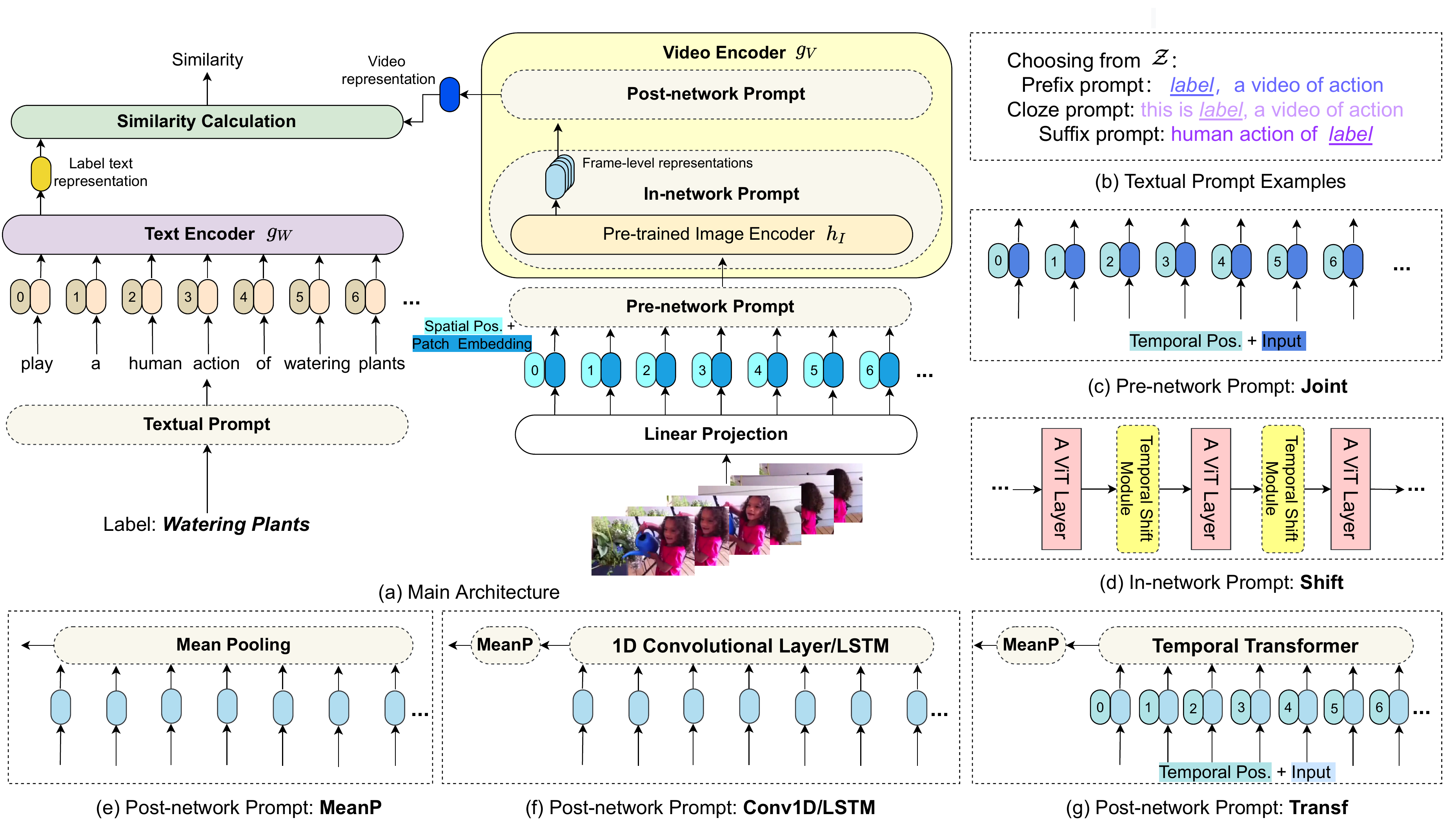}
	\caption{Overview of \textbf{ActionCLIP}. We present the whole architecture in (a), which consists of two single-modal encoders, a similarity calculation module and all possible prompt locations. (b) shows several examples of the textual prompt. (c) and (d) are the in-network and pre-network visual prompts details, respectively. (e),(f) and (g) give the details of post-network visual prompts, \textit{MeanP}, \textit{Conv1D}, \textit{LSTM} and \textit{Transf}. \textit{Pos.} is short for positional.}
	\label{fig:pipeline}
\end{figure*}
\subsection{The New Paradigm}
When considering the above multimodal learning framework, we need to consider the deficiency of label words. The most intuitive way is to take advantage of vast web image-text or video-text data. To cater for this, we propose a new ``\textit{pre-train, prompt and fine-tune}" paradigm for action recognition.\\
\noindent \textbf{Pre-train.} As prior arts suggested, pre-training has a large impact on vision-language multimodal learning~\cite{lu2019vilbert,lei2021less,li2021align,kim2021vilt}. Since the training data is directly collected from the web, one of the hot topics is to design appropriate objectives to handle these noisy data during this procedure. We find there are mainly three upstream pre-training proxy tasks in the pre-training procedure: multimodal matching (MM), multimodal contrastive learning (MCL) and masked language modeling (MLM). MM predicts whether a pair of modalities is matched or not. MCL aims to draw pairwise unimodal representations close to each other. MLM utilizes the features of both modalities to predict the masked words. However, this paper does not focus on this step due to the restriction of enormous computation cost. We directly choose to apply a pre-trained model and make efforts on the following two steps.\\
\noindent \textbf{Prompt.} Prompt in NLP means the original input is
modified using a template into a textual string prompt that has some unfilled slots to fill with expected results. Here we borrow the word ``\textit{prompt}" for the meaning of \textit{adjusting and reformulating the downstream tasks to act more like the upstream pre-training tasks}. Notably, the traditional practice is adapting the pre-trained model to the downstream classification task via attaching a new linear layer to the pre-trained feature extractor, which is reversed to ours. Here we make two kinds of prompts, textual prompt and visual prompt. The former is significant for label text extension. Given a label $\textbf{\emph{y}}$, we first define a set of permissible values $\mathcal{Z}$, then the prompted textual input ${\textbf{\emph{y}}}'$ is obtained by a filling function $f_{fill}({\textbf{\emph{y}}},\textbf{\emph{z}})$, where $\textbf{\emph{z}}\in \mathcal{Z}$. There are three varieties of $f_{fill}$, \textit{prefix prompt}, \textit{cloze prompt} and \textit{suffix prompt}. They are classified based on the filling locations. For visual prompt, its designation mainly depends on the pre-trained model. If the model is pre-trained on video-text data, it is almost no extra reformulation for the visual part since the model is already trained to output video representations. While if the model is pre-trained with image-text data, then we should empower the model to learn the important temporal relationship of videos. Formally, given a video $\textbf{\emph{x}}$, we introduce prompt function as $f_{tem}(h_I(x))$, where $h_I$ is the visual encoding network of pre-trained models. Similarly, $f_{tem}$ has three variants based on where it works against $h_I$, \textit{pre-network prompt}, \textit{in-network prompt} and \textit{post-network prompt}. With the elaborate designation of \textit{prompt}, we could even avoid the above unreachable computational ``\textit{pre-train}" step by keeping the learned ability of a pre-trained model. Note that in the new paradigm, the pre-trained model should not be largely modified due to \textit{catastrophic forgetting}~\cite{mccloskey1989catastrophic}, where the pre-trained model loses its ability to do things that it was able to do in the pre-training. We also demonstrate this point in our experiments.\\
\noindent \textbf{Fine-tune.} When there are sufficient downstream training datasets like Kinetics, it is no doubt that fine-tuning on specific datasets will dramatically improve the performance. Also, if the prompt introduces extra parameters, it is necessary to train these parameters and learn with the whole framework end-to-end. 

\subsection{New Paradigm Instantiation Details}\label{Instantiation}
Each component of the new paradigm has a wide variety of choices. As presented in Figure \ref{fig:pipeline}, we show an instantiation example here and conduct all the experiments with this instantiation. 

We employ a firsthand pre-trained model, CLIP~\cite{radford2021learning} to avoid the enormous computational resources at the first pre-training step. This instantiation model is called \textbf{ActionCLIP} as shown in Figure \ref{fig:pipeline}(a). CLIP is an efficient image-text representation trained with the MCL task, similar to our multimodal learning framework. Figure \ref{fig:pipeline}(b) shows concrete examples of the textual prompts used in the instantiation. We define $\mathcal{Z}$ to be $K$ discrete manual sentences which is the most natural way based on human introspection. Then the prompted input ${\textbf{\emph{y}}}'$ is fed into the language encoder $g_{W}$ that is the same with pre-trained language model $h_W$. For the vision model, based on the pre-trained image encoder $h_I$ of CLIP, we employ three types of visual prompts as follows.\\


\noindent \textbf{Pre-network Prompt.} This type operates on the inputs before feeding into the encoder, as shown in Figure \ref{fig:pipeline}(c). Given a video $\textbf{\emph{x}}$, we simply forward all spatio-temporal tokens extracted from the video through the visual encoder to jointly learn spatio-temporal attentions. Except for the spatial positional embedding, an extra learnable temporal positional embedding will be added to the token embedding to indicate the frame index. $g_{V}$ could use the original pre-trained image encoder $h_{I}$. We call this type \textit{Joint} for short.

\noindent \textbf{In-network Prompt.} We attempt a parameter-free prompt abbreviated as \textit{Shift} for this type as shown in Figure \ref{fig:pipeline}(d). We introduce the temporal shift module~\cite{lin2018temporal}, which shifts part of the feature channels along the temporal dimension and facilitates information exchanged among neighboring input frames. We insert the module between every two adjacent layers of $g_{V}$. The architecture and pre-trained weights of $g_{V}$ could directly reuse $h_{I}$ since this module brings no parameters. 

\noindent \textbf{Post-network prompt.} Given a video $\textbf{\emph{x}}$ with $F$ extracted frames, we sequentially encode spatial and temporal features with two separate encoders in this prompt. The first is a spatial encoder $g_{V}^{sp}$ which is responsible for only modeling interactions between tokens extracted from the same temporal index. We use $h_{I}$ as our $g_{V}^{sp}$. The extracted frame-level representations $u_{i}\in \mathbb{R}^{d}$ are then concatenated into $\textbf{u}\in \mathbb{R}^{F \times d}$, and then fed to a temporal encoder $g_{V}^{te}$ to model interactions between tokens from different temporal indices. We offer four choices for $g_{V}^{te}$, \textit{MeanP}, \textit{Conv1D}, \textit{LSTM} and \textit{Transf}, presented in Figure \ref{fig:pipeline}(e-g). \textit{MeanP} is short for mean pooling on the temporal dimension. \textit{Conv1D} is a 1d convolutional layer applied on the temporal dimension. \textit{LSTM} is a recurrent neural network and \textit{Transf} means a $L_{t}$ layer temporal vision transformer encoder. Since the temporal dimensions of \textit{Conv1D}, \textit{LSTM} and \textit{Transf} keep the same with input $\mathbf{u}$, the subsequent operations are the same as \textit{MeanP}.

Then we end-to-end fine-tune the whole network with the training objective Equation \ref{kl}. 
\section{Experiments}

\subsection{Experimental Setup} 
\noindent \textbf{Network architectures.} Our textual encoder $g_{W}$ follows that of CLIP which is a 12-layer, 512-wide Transformer with 8 attention heads and the activations from the highest layer at [\textrm{EOS}] are treated as the feature representation $\mathbf{w}$. We use ViT-B/32 and ViT-B/16 of CLIP's visual encoder $h_{I}$. They are all 12-layer vision transformers, with different input patch sizes of 32 and 16 respectively. The [\textrm{Class}] token of their highest layers' outputs are used. We use $K$=18 permissible values $\mathcal{Z}$ for textual prompt. For visual prompts, the layer of \textit{Conv1D} and \textit{LSTM} is 1, \textit{Transf} has $L_{t}$=6 layers. Two versions of \textit{Transf} are implemented, they are different in not using or using [\textrm{Class}] token. We distinguish them as \textit{Transf} and \textit{Transf$_{cls}$}. \\
\noindent \textbf{Training.} We use AdamW optimizer with a base learning rate of $5\times10^{-6}$ for pre-trained parameters and $5\times10^{-5}$ for new modules with learnable parameters. Models are trained with 50 epochs and the weight decay is 0.2. The learning rate is warmed up for 10\% of the total training epochs and decayed to zero following a cosine schedule for the rest of the training. The spatial resolution of the input frames is $224\times224$. We use the same segment-based input frame sampling strategy as \cite{wang2018temporal} with 8, 16 or 32 frames. Even the largest model of our method, ViT-B/16 could be trained with 4 NVIDIA GeForce RTX 3090 GPUs on Kinetics-400 when inputting 8 frames, and the training process takes about 2.5 days. Compared to X3D and SlowFast, both trained with 128 GPUs for 256 epochs, our training is much faster and requires fewer GPUs ($\sim$30).\\ 
\noindent \textbf{Inference.} The input resolution is $224 \times 224$ in all the experiments. Following \cite{jiang2019stm}, we use multi-view inference with 3 spatial crops and 10 temporal clips of each video only for the best performance model. The final prediction is from the averaged similarity scores of all views.

\subsection{Ablation Experiments}
In this section, we do extensive ablation experiments to demonstrate our method with the instantiation, \textbf{ActionCLIP}. Models in this section use 8-frame input, Transf for temporal modeling, ViT-B/32 as the backbone and single view testing on Kinetics-400, unless specified otherwise. 

\noindent \textbf{Is the ``\textit{multimodal framework}" helpful?} To compare with the traditional video-unimodal 1-of-N classification model, we implement a variant called \textit{unimodality} which has the same backbone, pre-trained weights and temporal modeling strategy (before the final linear layer) with our \textbf{ActionCLIP}. The results are shown in Table \ref{ab1}. When exploiting the semantic information of label texts with our multimodal learning framework, it dramatically improves the performance with 2.91\% top-1 accuracy gains, demonstrating that the multimodal framework is helpful to learn powerful representations for action recognition. 
	\begin{table}[!htb]
		\centering
		\caption{Ablation of the ``\textit{multimodal framework}".}
		\label{ab1}
		\begin{tabular}{c|c|c}
			\hline
			\rowcolor[HTML]{ECF4FF} 
			& Unimodality & \textbf{ActionCLIP}   \\ \hline
			Top-1 & 75.45 &  78.36   \\ \hline
			Top-5 &  92.51 &  94.25  \\ \hline
		\end{tabular}
	\end{table}

\noindent \textbf{Is the ``\textit{pre-train}" step important?} In Table \ref{ab2}, we validate the impact of this step by experimenting with random initialized or CLIP pre-trained vision and language encoders. In particular, from the large gap (40.10\% \textit{vs.} 78.36\%) between model V3 and V4, we find that the visual encoder needs proper initialization, otherwise the model will fail to obtain a strong performance. The language encoder has a smaller influence, since model V2 could also get a comparative result (76.63\%) compared with model V4 (78.36\%). When both the visual and language encoders are randomly initialized, model V1 is hard to learn a good representation and drops a large margin of 41.4\% from model V4. Therefore, the final conclusion is that the ``\textit{pre-train}" step is important, especially for the visual encoder. 

	\begin{table}[!htb]
		\centering
		\caption{Ablation of the ``\textit{pre-train}" step.}
		\label{ab2}
		\begin{tabular}{c|c|c|c|c}
			\hline
			\rowcolor[HTML]{ECF4FF} 
			Model & Language & Vision & Top1 & Top5 \\ \hline
			V1 & random & random & 36.96 & 63.02 \\ \hline
			V2 & random & CLIP & 76.63 & 91.94 \\ \hline
			V3 & CLIP & random & 40.10 & 66.35 \\ \hline
			V4 & CLIP & CLIP & 78.36 & 94.25 \\ \hline
		\end{tabular}
	\end{table}
	
	\begin{table}[!htb]
		\centering
		\caption{Ablation of the textual \textit{prompt}.}
		\label{ab3_1}
		\begin{tabular}{c|c|c}
			\hline
			\rowcolor[HTML]{ECF4FF} 
			& only label & textual prompt \\ \hline
			Top-1 & 77.82      & 78.36          \\ \hline
			Top-5 & 93.95      & 94.25   \\   \hline
		\end{tabular}
	\end{table}
\begin{figure*}[ht]
	\centering
	\begin{minipage}[b]{0.33\linewidth}
		\centering
		\centerline{
			\includegraphics[width=1.0\linewidth,height=0.78\linewidth]{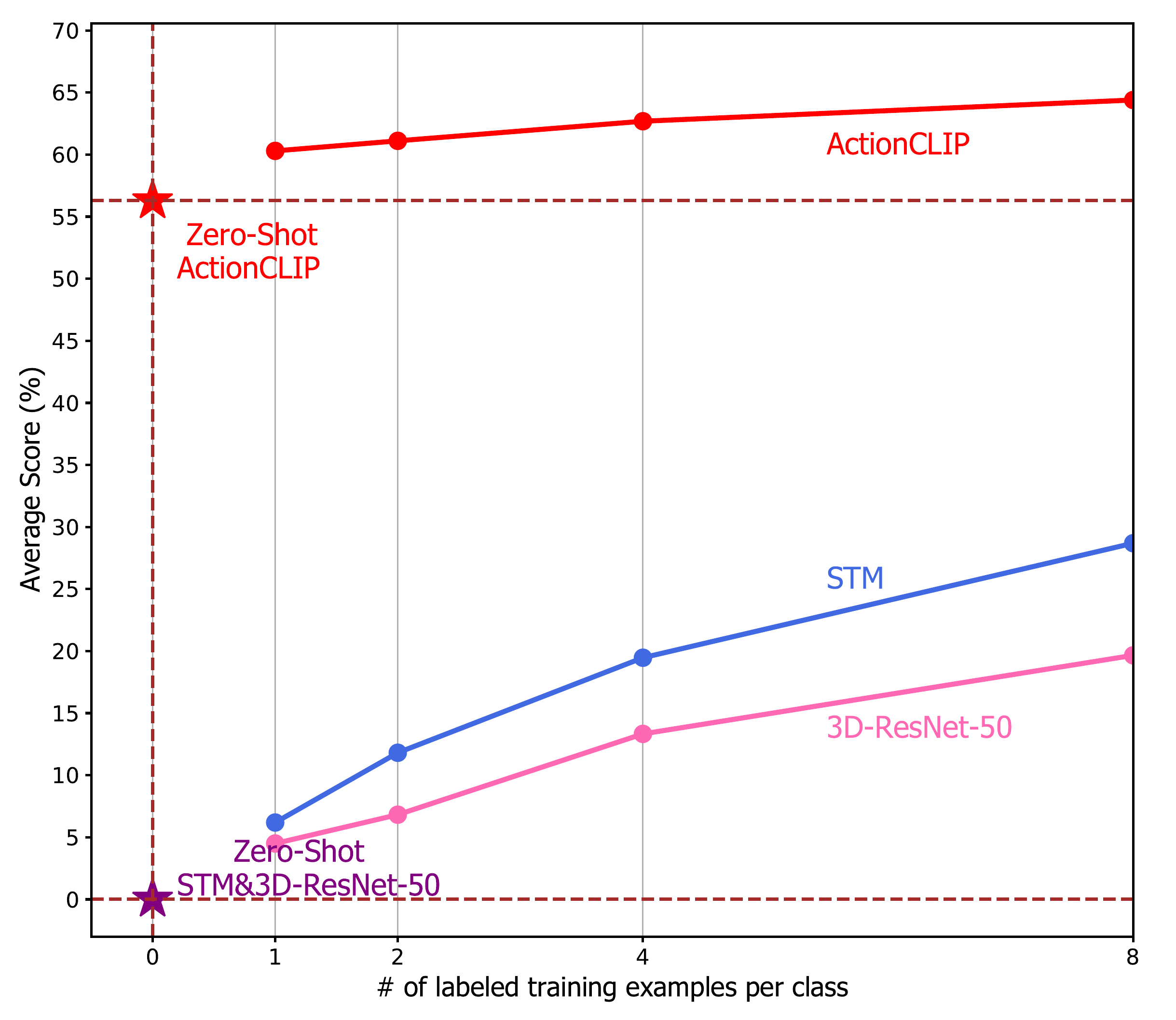}
			\includegraphics[width=1.0\linewidth,height=0.78\linewidth]{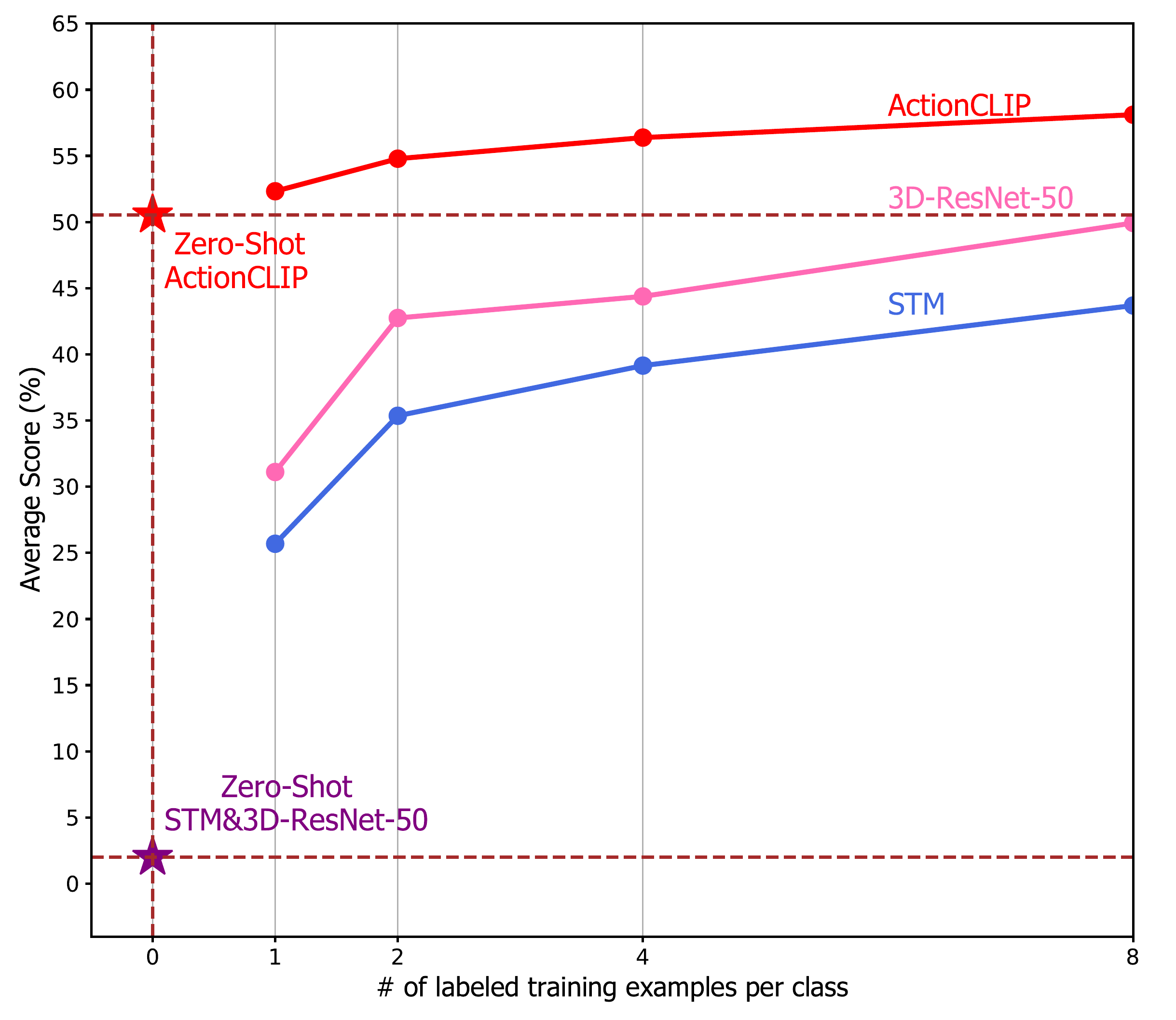}
			\includegraphics[width=1.0\linewidth,height=0.78\linewidth]{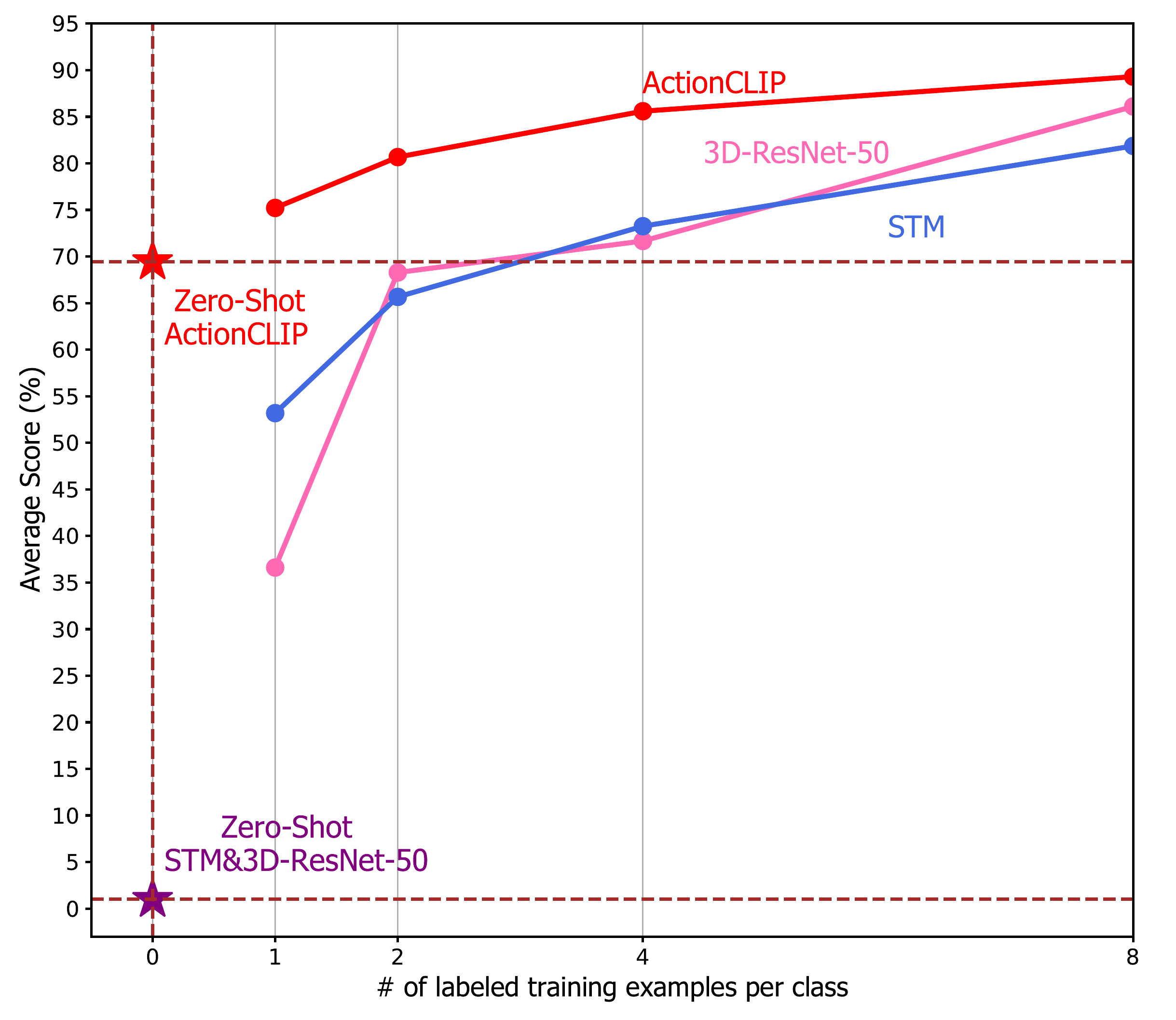}
		}
	\end{minipage}
	\caption{Zero-shot/few-shot results on Kinetics-400 (left), HMDB-51 (middle) and UCF-101 (right). \textbf{ActionCLIP} leads the performance under these hard data-poor circumstances. It can do zero-shot recognition on all three datasets while STM and 3D-ResNet-50 are not available under this condition. Also, \textbf{ActionCLIP} is good at few-shot classification where the performance gap is obvious compared to the other two methods. Brown dashed lines demonstrate zero-shot accuracy.}
	\label{fig:zeroshot}
\end{figure*}

\noindent \textbf{Is the ``\textit{prompt}" step important?} Table \ref{ab3_1} shows the results of textual prompt. It can be seen that using only the label words drops 0.54\% compared with using textual prompt, demonstrating the validness of this simple, discrete and human-comprehensible textual prompt. For the visual prompt, note that \textit{MeanP} is the simplest temporal fusion way and we compare other visual prompts with it. As shown in Table \ref{ab3_2}, we find \textit{Joint} and \textit{Shift} obviously decrease the performance by 2.74\% and 5.38\%, respectively. We believe the reason is the \textit{catastrophic forgetting} phenomenon since the input pattern is changed in \textit{Joint} and the features of pre-trained image encoder $h_{I}$ are changed in \textit{Shift}. These operations may break the original learned strong representations and yield performance drop. Post-network prompts are more suitable and safer options to keep the learned character. Specifically, \textit{LSTM} and \textit{Conv1D} cause a negligible top-1 drop but they all improve the top-5 accuracy. \textit{Transf$_{cls}$} and \textit{Transf} improve the top-1 results with 1.01\% and 1.25\%. We choose \textit{Transf} as our final visual prompt since it has the best results. In a word, the designation of \textit{prompt} is significant since proper prompts could avoid \textit{catastrophic forgetting} and maintain the representation power of existing pre-trained models, giving a shortcut to the usage of tremendous web data.
	\begin{table}[!htb]
		\centering
		\caption{Ablation of the visual \textit{prompt}.}
		\label{ab3_2}
		\begin{tabular}{c|c|c|c}
			\hline
			\rowcolor[HTML]{ECF4FF} 
			\multicolumn{2}{c|}{Visual prompt}         & Top-1 &Top-5 \\ \hline
			Pre-network                   & Joint       & 74.37 &93.09  \\ \hline
			In-network                    & Shift       & 71.73   & 91.07    \\ \hline
			\multirow{5}{*}{Post-network} & MeanP       & 77.11  & 93.79   \\ \cline{2-4} 
			& LSTM        & 77.09  &93.86         \\ \cline{2-4} 
			& Conv1D      & 77.04 &94.06       \\ \cline{2-4} 
			& Transf$_{cls}$ & 78.12  &94.25         \\ \cline{2-4} 
			& Transf      & 78.36  &94.25       \\ \hline
		\end{tabular}
	\end{table}

\noindent \textbf{Is the ``\textit{fine-tune}" step important?} We demonstrate this step by separately freezing the parameters of the pre-trained language encoder $h_{W}$ and image encoder $h_{I}$. The results are presents in Table \ref{ab4}. When the two encoders are all frozen and only the visual prompt \textit{Transf} is trained, the performance decreases 6.15\% on top-1 accuracy. When all the parameters are end-to-end fine-tuned, we obtain the best results of 78.36\%. It will have a negative influence on the accuracy if either of the pre-trained encoders is frozen. Therefore, the ``\textit{fine-tune}" step is indeed crucial to specific datasets, which is consistent with our perceptual intuition.  
	\begin{table}[h]
		\centering
		\caption{Is the ``\textit{fine-tune}" step important? ``$\checkmark$" means do fine-tuning while ``$\times$" means fixing the parameters without fine-tuning.} 
		\label{ab4}
		\begin{tabular}{c|c|c|c|c}
			\hline
			\rowcolor[HTML]{ECF4FF} 
			Model & Language & Video & Top-1 & Top-5 \\ \hline
			V1    & $\times$    & $\times$   & 72.21 & 91.61 \\ \hline
			V2    & $\times$    & $\checkmark$ & 73.88 & 91.61 \\ \hline
			V3    &$\checkmark$  &$\times$     & 73.78 & 92.17 \\ \hline
			V4    &$\checkmark$ &$\checkmark$ & 78.36 & 94.25 \\ \hline
		\end{tabular}
	\end{table}
\noindent \textbf{Backbones and input frames.} In Table \ref{ab5}, we experiment \textbf{ActionCLIP} with different backbones and input frames configurations. The input frames vary from 8, 16 to 32. Two different backbones are used, ViT-B/32 and ViT-B/16. The conclusion is intuitive that larger models and more input frames yield better performance. 
	\begin{table}[!htb]
		\centering 
		\caption{Influence of backbones and input frames.}
		\label{ab5}
		\begin{tabular}{c|c|c|c}
			\hline
			\rowcolor[HTML]{ECF4FF} 
			Backbone                & Input frames & Top1  & Top5  \\ \hline
			ResNet-50                 & 8            & 73.31     & 92.02     \\ \hline
			ViT-B/32                  & 8            & 78.36 & 94.25 \\ \hline
			\multirow{3}{*}{ViT-B/16} & 8            & 81.09 & 95.49 \\ \cline{2-4} 
			& 16           & 81.68 & 95.87 \\ \cline{2-4} 
			& 32           & 82.32 & 96.20  \\ \hline
		\end{tabular}
	\end{table}

\subsection{Runtime Analysis}
For different backbones and input frame configurations, we present their model sizes, FLOPs and inference speeds in Table \ref{runtime}. The textual encoder of all backbones has the same architecture, which has 37.8M parameters. We show the whole parameter in the table. Notably, ViT-B/32 has a little more parameters than ViT-B/16, which comes from the linear projection layer before feeding into the vision transformer. While ViT-B/32 has much a faster inference speed (3.3$\times$) and fewer FLOPs (4$\times$) than ViT-B/16. Moreover, we provide two very recent methods for comparison, TimeSformer~\cite{bertasius2021space} and ViViT~\cite{arnab2021vivit} with their highest configurations which obtain similar accuracy with \textbf{ActionCLIP}. Specifically, compared with the highest configuration of \textbf{ActionCLIP}, TimeSformer needs more input frames (3$\times$) and much more computational FLOPs (12.7$\times$) to obtain its best performance, which is still worse than \textbf{ActionCLIP} (82.3\% \textit{vs.} 80.7\%). Similarly, ViViT has much more computational FLOPs (7.1$\times$) to obtain its best results with a larger input solution 320$\times$320, while \textbf{ActionCLIP}'s input is 224$\times$224 and it surpasses ViViT with 1\% top-1 accuracy gap and runs faster (3.1$\times$)  than ViViT. In conclusion, \textbf{ActionCLIP} is a cost-effective and efficient method for action recognition.

\begin{table}[h]
	\centering
	\caption{Parameters, FLOPs and inference speed comparison. All the results of \textbf{ActionCLIP} are tested on one NVIDIA GeForce RTX 3090 GPU and use single-view testing.}
	\label{runtime}
	\setlength\tabcolsep{2pt}
	\scalebox{0.9}{\begin{tabular}{c|ccccc}
			\hline
			\rowcolor[HTML]{ECF4FF} 
			Backbone                   & Frames & Top-1 & GFLOPs & Params & Runtime  \\ \hline
			TimesSformer-L & 96     & 80.7 & 7140  & -                               & -  \\
			ViViT-L/16 320 & 32     & 81.3 & 3992  & -                               & 4.2V/s  \\ \hline
			ViT-B/32                   & 8      & 78.4 & 35.4   & 144.1M                               & 144.7V/s \\ \hline
			& 8      & 81.1 & 140.8  & 141.7M                               & 43.2V/s  \\ \cline{2-6} 
			& 16     & 81.7 & 281.6  & 141.7M                               & 21.2V/s  \\ \cline{2-6} 
			\multirow{-3}{*}{ViT-B/16} & 32     & \textbf{82.3} & 563.1  & 141.7M                               & 13.0V/s  \\ \hline

	\end{tabular}}
\end{table}

\subsection{Zero-shot/few-shot Recognition} 
In this section, we demonstrate the attractive zero-shot/few-shot recognition ability of our \textbf{ActionCLIP} (ViT-B/16). We implement two representative methods for comparison, STM~\cite{jiang2019stm} which is a well-designed temporal-encoded 2D network, and 3D-ResNet-50 which is the slow path of SlowFast~\cite{feichtenhofer2019slowfast}. We use 8-frame input, single view inference in all models of this section. We first conduct the zero-shot/few-shot experiments on Kinetics-400. \textbf{ActionCLIP} uses pre-trained model of CLIP with \textit{MeanP} visual prompt (since \textit{Transf} has no pre-trained parameters), STM and 3D-ResNet-50 are pre-trained on ImageNet. Then, we validate on UCF-101 and HMDB-51 with Kinetics-400 pre-trained models (\textbf{ActionCLIP} uses \textit{Transf} here). As shown in Figure \ref{fig:zeroshot}, the results demonstrate the strong transfer power of \textbf{ActionCLIP} under these data-poor conditions, while the traditional unimodality methods are not able to do zero-shot recognition and their few-shot performance is ineffective compared with \textbf{ActionCLIP} even pre-trained on large Kinetics-400.  
\begin{table}[t]
	\caption{Comparison with previous work on Kinetics-400. ``-" indicates the numbers are not available for us. ``+" in the second column means the different input of two paths of the method.} 
	\label{k400}
	\centering
	\scalebox{0.95}{\begin{tabular}{cccc}
			\hline
			\rowcolor[HTML]{FFCCC9}
			Methods                     & Frames        & Top-1 & Top-5 \\ \hline
			I3D NL \cite{wang2018non}               & 32      & 77.7  & 93.3  \\ 
			S3D-G \cite{xie2018rethinking}              & 64     & 74.7  & 93.4  \\ 
			SlowFast \cite{feichtenhofer2019slowfast}    & 16+64  & 79.8  & 93.9  \\ 
			X3D-XXL \cite{feichtenhofer2020x3d}             & 16      & 80.4  & 94.7  \\
			TPN \cite{yang2020temporal}                 & 32      & 78.9  & 93.9  \\
			SmallBigNet~\cite{li2020smallbignet}       & 32       & 77.4  & 93.3  \\
			CorrNet~\cite{wang2020video}            & 32      & 79.2  & -     \\ 
			\hline
			R(2+1)D~\cite{tran2018closer} & 32+32 & 75.4  & 91.9  \\ 
			TSM~\cite{lin2018temporal}               & 16      & 74.7  & -     \\ 
			TEA~\cite{li2020tea}               & 16      & 76.1  & 92.5  \\
			STFT~\cite{kumawat2021depthwise}              & 64      & 75.0    & 91.1  \\ 
			STM~\cite{jiang2019stm}                & 16      & 73.7  & 91.6  \\ 
			TANet~\cite{liu2020tam}              & 16       & 79.3  & 94.1  \\
			TEINet~\cite{liu2020teinet}             & 16     & 76.2  & 92.5  \\
			TDN~\cite{wang2021tdn}               & 8+16  & 79.4  & 93.9  \\ 
			\hline
			ViT-B-VTN~\cite{neimark2021video}          & 250            & 79.8  & 94.2  \\ 
			MViT-B~\cite{fan2021multiscale}             & 64      & 81.2  & 95.1  \\ 
			STAM \cite{sharir2021image}                                      & 64 & 80.5 & - \\
			TimeSformer-L \cite{bertasius2021space}                            & 96 & 80.7 & 94.7 \\
			ViViT-L/16x2~\cite{arnab2021vivit} & 32       & 80.6  & 94.7  \\ 
			ViViT-L/16x2 (JFT) & 32       & 82.8  & 95.3  \\ \hline
			\multirow{2}{*}{\textbf{ActionCLIP} (ViT-B/16)} & 16     & 82.6     & 96.2  \\ 
			& 32      & \textbf{83.8}     & \textbf{97.1}    \\ \hline
	\end{tabular}}
\end{table}

\subsection{Comparison with State-of-the-art Methods} 
In this section, we evaluate the performance of our method on a diverse set of action recognition datasets: Kinetics-400~\cite{carreira2017quo}, Charades~\cite{sigurdsson2016hollywood}, UCF-101~\cite{soomro2012ucf101} and HMDB-51~\cite{Kuehne11}. ViT-B/16, Transf prompt and multi-view testing are used in \textbf{ActionCLIP}. The results of UCF-101 and HMDB-51 are shown in Appendix.
\noindent \textbf{Kinetics-400.} Table \ref{k400} compares to prior methods on Kinetics-400. There are four parts in this table, corresponding to 3D-CNN-based methods, 2D-CNN-based methods, transformer-based methods and our method. According to the table, the third section achieves better results with strong vision transformers than the first and second parts. Among them, the first four methods build on the 12-layer ViT-B/16 model for parameter-accuracy balance, so do our \textbf{ActionCLIP}. ViViT instead uses a larger model, 24-layer ViT-L for better results. Also, it introduces JFT for pre-training for further gain. Our \textbf{ActionCLIP} achieves 82.6\% top-1 accuracy with only 16-frame input, which exceeds all the methods in the first and second parts of the table and most transformer-based methods that may use more input frames like 250 frames of ViT-B-VTN. An interesting discovery is that our top-5 accuracy is always higher than other methods.  We think this benefits from our multimodal framework's different inference process, which calculates the similarity between the semantic representations of videos and all labels. \textbf{ActionCLIP} further reaches a leading performance of 83.8\% when increasing the input to 32 frames. We believe that more input frames, larger models and larger input resolutions will yield better results and leave it to future work. The current performance of \textbf{ActionCLIP} could already reveal the potential of the multimodal learning framework and the proposed new paradigm for action recognition.


\begin{table}[pt] 
	\caption{Comparison with previous work on Charades. ``-" indicates the numbers are not available for us. ``+" in the second column means the different input of two paths of the method.}
	\label{charades}
	\centering
	\scalebox{0.95}{\begin{tabular}{ccc}
			\hline
			\rowcolor[HTML]{FFCCC9}
			Method                 & Frames & mAP \\ \hline
			MultiScale TRN \cite{zhou2018temporal}  & -   & 25.2 \\
			STM~\cite{jiang2019stm}     & 16   & 35.3   \\
			Nonlocal \cite{wang2018non}              & -   & 37.5   \\
			STGR+NL \cite{wang2018videos}            & -  & 39.7   \\
			SlowFast 50 \cite{feichtenhofer2019slowfast}      & 8+32   & 38.0   \\
			SlowFast 101+NL      & 16+64  & 42.5 \\
			X3D-XL (312) \cite{feichtenhofer2020x3d}   &16   & 43.4      \\
			LFB+NL \cite{wu2019long}           &32  &42.5  \\
			Timeception\cite{hussein2019timeception}                                          & - & 41.1\\ \hline
			\textbf{ActionCLIP} (ViT-B/16)     & 32      & \textbf{44.3}  \\ \hline
			
	\end{tabular}}
\end{table}

\noindent \textbf{Charades.} This is a dataset with longer-range activities and it has multiple actions inside every video. We show Kinetics-400 pre-trained models in Table \ref{charades}. Mean Average Precision (mAP) is used for evaluation. \textbf{ActionCLIP} achieves the top performance of 44.3 mAP, which demonstrates its effectiveness on multi-label video classification.

\section{Conclusion}
This paper provides a new perspective for action recognition by regarding it as a video-text multimodal learning problem. Unlike the canonical approaches that model the task as a video unimodality classification problem, we propose a multimodal learning framework to exploit the semantic information of label texts. Then, we formulate a new paradigm, i.e., ``\textit{pre-train, prompt, and fine-tune}" to enable our framework to directly reuse powerful large-scale web data pre-trained models, greatly reducing the pre-training cost. We implement an instantiation of the new paradigm, \textbf{ActionCLIP}, which has a superior performance on both general and zero-shot/few-shot action recognition. We hope our work could provide a new perspective for this task, especially raising attention on language modeling.
\section{Acknowledgements}
We would like to thank Zeyi Huang for his constructive suggestions and comments on this work.
{\small
	\bibliographystyle{ieee_fullname}
	\bibliography{aaai22.bib}
}


\end{document}